\documentclass{article}

\usepackage{microtype}
\usepackage{graphicx}
\usepackage{subfigure}
\usepackage{amsmath}
\usepackage{amssymb}
\usepackage{todonotes}
\usepackage{booktabs} 

\usepackage[draft]{hyperref}

\usepackage[accepted]{icml2019}

\icmltitlerunning{}

\begin{document}

\twocolumn[
\icmltitle{Learning Compact Neural Networks Using \\Ordinary Differential Equations as Activation Functions}




\begin{icmlauthorlist}
\icmlauthor{MohamadAli Torkamani}{amazon,uo}
\icmlauthor{Phillip Wallis}{microsoft}
\icmlauthor{Shiv Shankar}{umass}
\icmlauthor{Amirmohammad Rooshenas}{umass}
\end{icmlauthorlist}

\icmlaffiliation{amazon}{Amazon Inc.}
\icmlaffiliation{microsoft}{Microsoft Inc.}

\icmlaffiliation{umass}{University of Massachusetts Amherst}
\icmlaffiliation{uo}{Most of the work has been done when the author was affiliated with the University of Oregon.}
\icmlcorrespondingauthor{Mohammadali Torkamani}{alitor@amazon.com}


\vskip 0.3in
]


\printAffiliationsAndNotice{}  

\begin{abstract}
Most deep neural networks use simple, fixed activation functions, such
as sigmoids or rectified linear units, regardless of domain or
network structure. We introduce differential equation units (DEUs), an
improvement to modern neural networks, which enables each neuron to learn a particular nonlinear activation function from a family of solutions to an ordinary differential equation. Specifically, each neuron may change its functional form during training based on the behavior of the other parts of the network.
We show that using neurons with DEU activation functions results in a more compact network capable of achieving comparable, if not superior, performance when is compared to much larger
networks.
\end{abstract}

\section{Introduction}
Driven in large part by advancements in storage, processing, and parallel computing, deep neural networks (DNNs) have become capable of outperforming other methods across a wide range of highly complex tasks. Although DNNs often produce better results than shallow methods from a performance perspective, one of the main drawbacks of DNNs in practice is computational expense. One could attribute much of the success of deep learning in recent years to cloud computing and GPU processing. While deep learning based applications continue to be integrated into all aspects of modern life, future advancements will continue to be dependent on the ability to perform more operations, faster, and in parallel unless we make fundamental changes to the way these systems learn. 

State-of-the-art DNNs for computer vision, speech recognition, and natural language processing require too much memory, computation, and power to be run on current mobile or wearable devices. To run such applications on mobile, or other resource-constrained devices, either we need to use these devices as terminals and rely on cloud resources to do the heavy lifting, or we have to find a way to make DNNs more compact. For example, ProjectionNet~\cite{ravi2017projectionnet} and MobileNet~\citep{howard2017mobilenets} are both examples of methods that use compact DNN representations with the goal of on-device applications. In ProjectionNet, a compact \textit{projection} network is trained in parallel to the primary network, and is used for the on-device network tasks. MobileNet, on the other hand, proposes a streamlined architecture in order to achieve network compactness. One drawback to these approaches is that network compactness is achieved at the expense of performance. In this paper, we propose a different method for learning compact, powerful, stand-alone networks: we allow each neuron to learn its individual activation function enabling a compact neural network to achieve higher performance.

We introduce differential equation units (DEUs) where the activation function of each neuron is the nonlinear, possibly periodic solution of a second order, linear, ordinary differential equation. 
From an applicability perspective, our approach is similar to max-out networks \citep{goodfellow2013maxout}, adaptive piece-wise linear units (PLUs)~\cite{agostinelli2014learning,ramachandran2017searching}.
While the number of parameters learned by max-out and PLU is proportional to the number of input weights to a neuron, and the number of linear units in that neuron, for each DEU we learn only five additional parameters. Moreover, different from aforementioned activation functions, DEUs transform themselves during network training, and different neurons may utilize different forms for their activation functions. This variety of forms throughout a network enable it to encode more information, thus requiring less neurons for achieving the same performance comparing to the networks with fixed activation functions.


The advent of new activation functions such as rectified linear units (ReLU) \citep{nair2010rectified}, exponential linear units (ELU) \citep{clevert2015fast}, and scaled exponential linear units (SELU) \citep{klambauer2017self} address a networks ability to effectively learn complicated functions, thereby allowing them to perform better on complicated tasks. The choice of an activation function is typically determined empirically by tuning, or due to necessity. For example, in modern deep networks, ReLU activation functions are often favored over  sigmoid functions, which used to be a popular choice in the earlier days of neural networks. A reason for this preference is that the ReLU function is non-saturating and does not have the vanishing gradient problem when used in deep structures \citep{hochreiter1998vanishing}. 




Our contributions in this paper include the following: We introduce differential equation units. We propose a learning process to learn the parameters of a differential equation for each neuron. We empirically show that neural networks with DEUs can achieve high performance with more compact representations and are effective for solving real-world problems.

\section{Differential Equation Units}
Inspired by functional analysis and calculus of variations~\cite{gel1963variatsionnoeischislenie,gelfand2000calculus}, instead of using a fixed activation function for each layer, we propose a novel solution for learning an activation function for each neuron in the network. 


The main idea is to find the parameters of an ordinary differential equation (ODE) for each neuron in the network, whose solution would be used as the activation function of the neuron. As a result, each neuron learns a personalized activation function flexibly. We select (learn) the parameters of the differential equation from a low dimensional space (i.e., five). By minimizing the network loss function, our learning algorithm smoothly updates the parameters of the ODE, which results in an uncountably\footnote{Up to computational precision limitations.} extensive range of possible activation functions.

We parameterize the activation function of each neuron using a linear, second order ordinary differential equation $a y''(t) + b y'(t) + c y(t) = u(t)$, parameterized by five coefficients ($a$, $b$, $c$, $c_1$, $c_2$), where $a$, $b$, and $c$ are the scalars that we use to parameterize the ODE, $c_1$ and $c_2$ represent the initial conditions of the ODE's solution, and $u(t)$ is a regulatory function that we call the \textit{core activation function}. The coefficients are the only additional parameters that we learn for each neuron and are trained by the backpropagation algorithm.  To simplify the math and because it is a standard practice in control theory, we have set $u(t)$ to the Heaviside step function: $u(t) = 1$ for $x>0$ and $0$ otherwise.

In engineering and physics, such a model is often used to denote the exchange of energy between mass and stiffness elements in a mechanical system or between capacitors and inductors in an electrical system~\cite{ogata2002modern}. Interestingly, by using the solutions of this formulation as activation functions, we can gain a few key properties: approximation or reduction to some of the standard activation functions such as sigmoid or ReLU; the ability to capture oscillatory forms; and, exponential decay or growth.

\begin{table*}[h]
\centering
\begin{tabular}{l r || r r}
\hline
\toprule
Model & Size & MNIST & Fashion-MNIST \\

\toprule
MLP-ReLU & 1411k & 98.1 &  89.0\\ 
CNN-ReLU & 30k &99.2 & \textbf{90.2} \\
\hline
MLP-DEU & 1292k & 98.3 &   89.8 \\
CNN-DEU & 21k & \textbf{99.2} & 89.7  \\
\hline
Logistic Circuit & 460k & 97.4 & 87.6 \\
\bottomrule
\end{tabular}
\caption{\label{tab:exp_img} Test accuracy of different models on the MNIST and Fashion-MNIST image classification task.}
\end{table*}

\begin{table*}
\centering
\begin{tabular}{l r ||r r r r}
\toprule
Architecture & Size & ReLU & PReLU&  Swish & DEU  \\
\toprule
ResNet-18 &  11174k & 91.25 & 92.1 &91.9 & \textbf{92.5} \\
Preact-ResNet & 11170k & 92.1 & 92.2 & 92.0 & \textbf{92.3}  \\
ResNet-Stunted & 678k & 89.3 & 89.4 &90.1 & \textbf{90.7} \\
\bottomrule
\end{tabular}
\caption{\label{tab:exp_cifar} Test accuracy using different ResNet architectures and activation functions on the CIFAR-10 image classification task.}
\end{table*}

\subsection{Learning Algorithm}
For fixed $a$, $b$ and $c$, the solution of the differential equation
will be $y = f(t;a,b,c) + c_1 f_1(t;a,b,c)+ c_2 f_2(t;a,b,c)$ for some functions $f$, $f_1$, $f_2$. $y$ lies on an affine space parameterized by scalars $c_1$ and $c_2$ that represent the initial conditions of the solution. Our learning algorithm has two main parts: solving the differential equations once, and using a backpropagation-based algorithm for jointly learning the network weights and the five parameters of each neuron.

First, we solve the differential equations parametrically and take the derivatives of the closed-form solutions: $\frac{\partial y}{\partial t}$ with respect to its input $t$, and $\frac{\partial y}{\partial a}$, $\frac{\partial y}{\partial b}$, $\frac{\partial y}{\partial c}$ with respect to parameter $a$, $b$, $c$. Moreover, the derivative with respect to $c_1$ and $c_2$ will be $f_1$ and $f_2$, respectively.

We use the backpropagation algorithm to update the values of DEU parameters $a$, $b$, $c$, $c_1$, and $c_2$ for each neuron along with using $\frac{\partial y}{\partial t}$ for updating network parameters $w$ (input weights to the neuron) and propagating the error to lower layers.


We initialize parameters  $a$, $b$, and $c$ for all neurons with a random positive number less than one and strictly greater than zero, while initializing $c_1=c_2=0.0$. Both neural networks parameters and DEU parameters are learned using
the conventional backpropagation algorithm with Adam updates~\citep{kingma2014adam}.


If one or two of the coefficients $a$, $b$, or $c$ are zero, then the solution of the differential equation falls into a singularity subspace that is different from the affine function space of neighboring positive or negative values for those coefficients. For example, for $b=0$ and $a*c>0$ , the solution will be $y(t) = \sin \left( {\frac {\sqrt {c}t}{\sqrt {a}}} \right) c_2+\cos
 \left( {\frac {\sqrt {c}t}{\sqrt {a}}} \right) c_1-{\frac {u(t) }{c} \left( \cos \left( {\frac {\sqrt 
{c}t}{\sqrt {a}}} \right) -1 \right) }$, but for $b=c=0$, we will have $y(t)=1/2\,{\frac {u(t) {t}^{2}}{a}}+c_1t+c_2$. We observe that changing $c>0$ to $c=0$ will change the resulting activation function from a pure ocsillatory form to a (parametric) leaky rectified
quadratic activation function. Our learning algorithm allows an activation function to jump over the singularity subspaces. However, if it falls into a singular subspace, the derivative with respect to the parameter has become zero and remains zero for the rest of training. Therefore, the training algorithm will continue to search for a better activation function only within the singular subspace.

In practice, for some hyperparameter $\epsilon$, if any one of $a$, $b$, or $c$ is less than $\epsilon$, we project that value to exactly zero, and use the corresponding solution from the singular sub-space. We do not allow $a=b=c=0$, and for this rare case we force $c = \epsilon$. During the learning process at most two of $a$, $b$, and $c$ can be zero, which creates seven possible subspaces (with $a, b, c \in \{\mathbb{R}-\{0\},\{0\}\}$) that are individually solved. 
Similarly, when $b^2-4ac$ is close to zero, the generic solution will be exponentially large, therefore if $-\epsilon<b^2-4ac<\epsilon$, we explicitly set $b = \sqrt(4ac)$ to stabilize the solution and to avoid large function values.

During training, we treat $a, b, c, c_1$, and $c_2$ like biases to the neuron (i.e., with input weight of $1.0$)
and update their values based on the direction of the corresponding
gradients in each mini-batch. The resulting activation functions can be highly nonlinear and may potentially involve exponential sub-components. Therefore large magnitude inputs to such neurons can lead to blowing up the inputs to next layers. In order to resolve this issue and to stabilize the network, we deploy a batch normalization method and separate the learning rate of network parameters and DEU parameters.

\section{Experiments}
To implement the DEUs, we solved the differential equations and took their derivatives using the Maple software package~\citep{maplesoft}. Maple also generates optimized code for the solutions, by breaking down equations in order to reuse computations. Although we used Maple here, this task could have been done simply by pen and paper (although more time consuming). Since each DEU in a layer learns its particular activation function, we parallelize the computations of a layer activation function for the participating neurons over a GPU to achieve scalable performance in our implementation.
We evaluate DEU on different models considering the classification performance and model size. We first use MNIST and Fashion-MNIST as our datasets to assess the behavior of DEUs with respect to the commonly used ReLU activation function (Table \ref{tab:exp_img}). 
We have also added comparison with the recent logistic circuits \cite{LiangUDL18}, which learns a compact discriminative representation of a posterior distribution over the class variable. 
DEU are competitive or better than normal networks for these tasks while having substantially smaller number of parameters.

Next we perform a more direct comparison of the effect of DEU on classification performance against ReLU, PReLU~\cite{he2015delving}, and Swish~\cite{ramachandran2017searching} activation functions on the CIFAR-10 dataset. PReLU is similar to ReLU with a parametric leakage and Swish has the form of $f(x) = x*\textit{sigmoid}(\beta x)$ with a learnable parameter $\beta$.

For these experiments we kept the network architecture fixed to ResNet-18 \cite{he2016deep} and used the hyperparameter settings as in He et al.~(\citeyear{he2016deep}). We observe that DEUs gain more than \textbf{1}\% improvement in accuracy. We further show that this improvement persists across other model designs. First we use a preactivation ResNet \cite{he16preact}, which is a ResNet-like architecture with a slightly smaller size. Second, to assess suitability for reducing the model size, we experiment with a stunted ResNet-18, which is a standard ResNet-18 model with half of its blocks removed. The result of this comparison is presented in Table \ref{tab:exp_cifar}, which indicates that DEUs are constantly work better than the other activation functions. Moreover, using DEUs partially fills the performance gap between ResNet-18 and stunted ResNet, which suggests the usefulness of DEUs in training compact neural networks.



\newpage

\bibliography{all}

\begin{thebibliography}{18}
\providecommand{\natexlab}[1]{#1}
\providecommand{\url}[1]{\texttt{#1}}
\expandafter\ifx\csname urlstyle\endcsname\relax
  \providecommand{\doi}[1]{doi: #1}\else
  \providecommand{\doi}{doi: \begingroup \urlstyle{rm}\Url}\fi

\bibitem[Agostinelli et~al.(2014)Agostinelli, Hoffman, Sadowski, and
  Baldi]{agostinelli2014learning}
Agostinelli, F., Hoffman, M., Sadowski, P., and Baldi, P.
\newblock Learning activation functions to improve deep neural networks.
\newblock \emph{arXiv preprint arXiv:1412.6830}, 2014.

\bibitem[Clevert et~al.(2015)Clevert, Unterthiner, and
  Hochreiter]{clevert2015fast}
Clevert, D.-A., Unterthiner, T., and Hochreiter, S.
\newblock Fast and accurate deep network learning by exponential linear units
  (elus).
\newblock \emph{arXiv preprint arXiv:1511.07289}, 2015.

\bibitem[Gelfand \& Fomin(1963)Gelfand and
  Fomin]{gel1963variatsionnoeischislenie}
Gelfand, I.~M. and Fomin, S.
\newblock Variatsionnoeischislenie, fizmatgiz, moscow 1961. mr 28\# 3352.
  translation: Calculus of variations, 1963.

\bibitem[Gelfand et~al.(2000)Gelfand, Silverman, et~al.]{gelfand2000calculus}
Gelfand, I.~M., Silverman, R.~A., et~al.
\newblock \emph{Calculus of variations}.
\newblock Courier Corporation, 2000.

\bibitem[Goodfellow et~al.(2013)Goodfellow, Warde-Farley, Mirza, Courville, and
  Bengio]{goodfellow2013maxout}
Goodfellow, I.~J., Warde-Farley, D., Mirza, M., Courville, A., and Bengio, Y.
\newblock Maxout networks.
\newblock \emph{In Proceedings of the 30th International Conference on Machine
  Learning}, pp.\  1319–1327, 2013.

\bibitem[He et~al.(2015)He, Zhang, Ren, and Sun]{he2015delving}
He, K., Zhang, X., Ren, S., and Sun, J.
\newblock Delving deep into rectifiers: Surpassing human-level performance on
  imagenet classification.
\newblock In \emph{Proceedings of the IEEE international conference on computer
  vision}, pp.\  1026--1034, 2015.

\bibitem[He et~al.(2016{\natexlab{a}})He, Zhang, Ren, and Sun]{he16preact}
He, K., Zhang, X., Ren, S., and Sun, J.
\newblock Identity mappings in deep residual networks.
\newblock \emph{CoRR}, abs/1603.05027, 2016{\natexlab{a}}.

\bibitem[He et~al.(2016{\natexlab{b}})He, Zhang, Ren, and Sun]{he2016deep}
He, K., Zhang, X., Ren, S., and Sun, J.
\newblock Deep residual learning for image recognition.
\newblock In \emph{Proceedings of the IEEE conference on computer vision and
  pattern recognition}, pp.\  770--778, 2016{\natexlab{b}}.

\bibitem[Hochreiter(1998)]{hochreiter1998vanishing}
Hochreiter, S.
\newblock The vanishing gradient problem during learning recurrent neural nets
  and problem solutions.
\newblock \emph{International Journal of Uncertainty, Fuzziness and
  Knowledge-Based Systems}, 6\penalty0 (02):\penalty0 107--116, 1998.

\bibitem[Howard et~al.(2017)Howard, Zhu, Chen, Kalenichenko, Wang, Weyand,
  Andreetto, and Adam]{howard2017mobilenets}
Howard, A.~G., Zhu, M., Chen, B., Kalenichenko, D., Wang, W., Weyand, T.,
  Andreetto, M., and Adam, H.
\newblock Mobilenets: Efficient convolutional neural networks for mobile vision
  applications.
\newblock \emph{arXiv preprint arXiv:1704.04861}, 2017.

\bibitem[Kingma \& Ba(2014)Kingma and Ba]{kingma2014adam}
Kingma, D.~P. and Ba, J.
\newblock Adam: A method for stochastic optimization.
\newblock \emph{arXiv preprint arXiv:1412.6980}, 2014.

\bibitem[Klambauer et~al.(2017)Klambauer, Unterthiner, Mayr, and
  Hochreiter]{klambauer2017self}
Klambauer, G., Unterthiner, T., Mayr, A., and Hochreiter, S.
\newblock Self-normalizing neural networks.
\newblock \emph{arXiv preprint arXiv:1706.02515}, 2017.

\bibitem[Liang \& Van~den Broeck(2018)Liang and Van~den Broeck]{LiangUDL18}
Liang, Y. and Van~den Broeck, G.
\newblock Learning logistic circuits.
\newblock In \emph{Proceedings of the UAI 2018 Workshop: Uncertainty in Deep
  Learning}, 2018.

\bibitem[{Maple 2018}()]{maplesoft}
{Maple 2018}.
\newblock Maplesoft, a division of waterloo maple inc.
\newblock URL \url{https://www.maplesoft.com/}.

\bibitem[Nair \& Hinton(2010)Nair and Hinton]{nair2010rectified}
Nair, V. and Hinton, G.~E.
\newblock Rectified linear units improve restricted boltzmann machines.
\newblock In \emph{Proceedings of the 27th international conference on machine
  learning (ICML-10)}, pp.\  807--814, 2010.

\bibitem[Ogata \& Yang(2002)Ogata and Yang]{ogata2002modern}
Ogata, K. and Yang, Y.
\newblock \emph{Modern control engineering}, volume~4.
\newblock Prentice hall India, 2002.

\bibitem[Ramachandran et~al.(2017)Ramachandran, Zoph, and
  Le]{ramachandran2017searching}
Ramachandran, P., Zoph, B., and Le, Q.
\newblock Searching for activation functions.
\newblock \emph{arXiv preprint arXiv:1710.05941}, 2017.

\bibitem[Ravi(2017)]{ravi2017projectionnet}
Ravi, S.
\newblock Projectionnet: Learning efficient on-device deep networks using
  neural projections.
\newblock \emph{arXiv preprint arXiv:1708.00630}, 2017.

\end{thebibliography}
\bibliographystyle{icml2019}

\end{document}